\documentclass[letterpaper, 10pt, conference]{ieeeconf}

\IEEEoverridecommandlockouts%
\usepackage{times}
\usepackage[bookmarks=true]{hyperref}
\usepackage{xcolor}
\usepackage[utf8]{inputenc}
\usepackage[T1]{fontenc}
\usepackage{amsmath,amssymb,mathtools}
\usepackage{booktabs}
\usepackage{array}
\usepackage{graphicx}
\let\labelindent\relax
\usepackage{enumitem}
\usepackage{caption}
\usepackage{subcaption}
\usepackage{overpic}
\usepackage{tikz}
\usetikzlibrary{arrows.meta,positioning}
\usepackage{cleveref}
\crefformat{section}{\S#2#1#3}
\crefformat{subsection}{\S#2#1#3}

\DeclareMathOperator{\clip}{clip}
\DeclareMathOperator*{\argmax}{arg\,max}
\DeclareMathOperator{\sg}{sg}
\newcommand{\R}{\mathbb{R}}

\definecolor{csensed}{HTML}{CFE8FF}
\definecolor{csensedline}{HTML}{1971C2}
\definecolor{cpriv}{HTML}{FFE8CC}
\definecolor{cprivline}{HTML}{E8590C}
\definecolor{cfrozen}{HTML}{E9ECEF}
\definecolor{cfrozenline}{HTML}{495057}
\definecolor{clearned}{HTML}{D3F9D8}
\definecolor{clearnedline}{HTML}{2F9E44}
\definecolor{coutput}{HTML}{99E9F2}
\definecolor{coutputline}{HTML}{0C8599}
\definecolor{cpanelA}{HTML}{F3EFFA}
\definecolor{cpanelAline}{HTML}{8A6FB0}
\definecolor{cpanelB}{HTML}{F2F9F3}
\definecolor{cpanelBline}{HTML}{7EC98D}
\definecolor{cpanelC}{HTML}{EEF9FB}
\definecolor{cpanelCline}{HTML}{7FC7D3}

\newif\ifdraftmode\draftmodefalse
\ifdraftmode
  
  \newcommand{\TODO}[1]{\textcolor{red}{\textbf{[TODO: #1]}}}
\else
  
  \newcommand{\TODO}[1]{}
\fi

\begin{document}

\title{\LARGE \bf Terrain Traversal Without Motion Priors: Expert-Routed
  Humanoid Locomotion at Deployment Fidelity}

\author{Thanh Ly$^{1*}$, Chien Le$^{1*}$, Truong-Duy Dang$^{1*}$,\\
  Duy M.~H.~Nguyen$^{1,4,5,6}$, Duy-Dung Le$^{2}$, Vien Anh Ngo$^{1,2}$, An T.~Le$^{1,2,3}$%
  \thanks{$^{1}$VinRobotics, Vietnam.
    $^{2}$Center for AI Research (CAIR), VinUniversity, Vietnam.
    $^{3}$Intelligent Autonomous Systems, TU Darmstadt, Germany.
    $^{4}$German Research Center for Artificial Intelligence (DFKI), Germany.
    $^{5}$University of Stuttgart, Germany.
    $^{6}$International Max Planck Research School for Intelligent Systems (IMPRS-IS), Germany.}%
  \thanks{$^{*}$Equal contribution. Corresponding author: An T.~Le, \texttt{an@robot-learning.de}.}}

\maketitle
\thispagestyle{empty}
\pagestyle{empty}
\IEEEpeerreviewmaketitle%
\bstctlcite{IEEEexample:BSTcontrol}%

\begin{abstract}
  State-of-the-art perceptive humanoid locomotion is mostly built on retargeted motion capture, a
  generative motion prior, and a distillation stage. We report a complete system
  that does not require this pipeline for a humanoid to cross varied terrain.
  In its place, we provide a five-expert mixture routed by a terrain classifier over a sensed elevation grid,
  and target every foothold with a model-based planner, refined by six closed-form-gradient steps on an interpolated
  terrain-cost field instead of picked from a fixed grid. Driven through the same
  perception-to-actuation software a robot would run, inside a second, deployment-fidelity
  simulator, it crosses an $85$\,m five-segment waypoint course $18$ of $20$ times end-to-end
  ($90\%$, Wilson $95\%$ CI $70$--$97$) and succeeds $20/20$ on four of five segments scored
  alone, where an ablated configuration without the mixture and without refinement reaches $0\%$
  on both the stair and the hurdle segment. The continuous refinement moves $72.4\%$ of foot placements across at least one grid-cell boundary,
  which produces real sub-grid correction.
  We also ran matched component studies to report what does not account for it: one expert
  equals five on the training distribution, four label-free routers collapse to near-uniform
  activation, and a self-supervised terrain embedding loses to the privileged label.
\end{abstract}

\section{Introduction}\label{sec:intro}

\begin{figure*}[t]
  \centering
  \resizebox{\textwidth}{8.5cm}{%
    \begin{tikzpicture}[
        font=\small, align=center,
        sensed/.style={draw=csensedline, fill=csensed, rounded corners=2pt,
            inner sep=2pt, line width=0.6pt},
        priv/.style={draw=cprivline, fill=cpriv, rounded corners=2pt,
            inner sep=2pt, line width=0.6pt, dashed},
        frozen/.style={draw=cfrozenline, fill=cfrozen, rounded corners=2pt,
            inner sep=2pt, line width=0.6pt, dashed},
        learned/.style={draw=clearnedline, fill=clearned, rounded corners=2pt,
            inner sep=2pt, line width=0.6pt},
        outnode/.style={draw=coutputline, fill=coutput, rounded corners=2pt,
            inner sep=2pt, line width=0.6pt},
        arr/.style={-{Latex[length=1.3mm]}, line width=0.5pt},
        oarr/.style={arr, dashed, draw=cprivline},
        elab/.style={font=\small\itshape, fill=white, inner sep=1pt},
      ]
      \draw[fill=cpanelA, draw=cpanelAline, line width=0.8pt, rounded corners=4pt]
      (0.0,0.0) rectangle (17.5,-3.0);
      \node[fill=cpanelAline, text=white, font=\scriptsize\bfseries, anchor=north west,
        rounded corners=2pt, inner sep=3pt] at (0.15,-0.2) {TRAINING ONLY --- ABOVE THE DEPLOYMENT BOUNDARY};
      \draw[fill=cpanelB, draw=cpanelBline, line width=0.8pt, rounded corners=4pt]
      (0.0,-3.3) rectangle (12.75,-6.0);
      \node[fill=cpanelBline, text=white, font=\scriptsize\bfseries, anchor=north west,
        rounded corners=2pt, inner sep=3pt] at (0.15,-3.5) {MoE ACTOR / GATE};
      \draw[fill=cpanelC, draw=cpanelCline, line width=0.8pt, rounded corners=4pt]
      (0.0,-6.25) rectangle (12.75,-9.75);
      \node[fill=cpanelCline, text=white, font=\scriptsize\bfseries, anchor=north west,
        rounded corners=2pt, inner sep=3pt] at (0.15,-6.45) {LOCAL FOOTHOLD REFINEMENT};

      \node[sensed, text width=1.5cm] (depth) at (1.0,-1.5) {Onboard depth stream};
      \node[frozen, text width=2.1cm] (enc) at (3.5,-1.5) {Contrastive encoder $f_\psi$\\
        InfoNCE};
      \node[frozen, text width=1.4cm] (at) at (6.0,-1.5) {$a_t\in\R^{64}$};
      \node[priv, text width=1.9cm] (ylabel) at (8.0,-2.25) {Privileged label $y_t$\\ $K{=}5$ classes};
      \node[priv, text width=2.3cm] (critsig) at (12,-1.5)
      {Critic conditioner:\\ A: $a_t$ (self-sup.)\\ B: one-hot $y_t$ (priv.)};
      \node[priv, text width=2.1cm] (critpriv) at (15.5,-0.75)
      {Privileged critic-only: full observation};
      \node[priv, text width=2.0cm] (critic) at (15.5,-2.25)
      {Critic value head};

      \node[sensed, text width=1.5cm] (elev) at (1.0,-5.25)
      {Elevation grid $\mathcal{H}_t$};
      \node[learned, text width=1.5cm] (gate) at (3.0,-5.25) {Gate $g_\phi$\\ CNN, $K{=}5$};
      \node[learned, text width=1.5cm] (logits) at (5.0,-5.25) {Logits $\boldsymbol{\ell}_t$\\ over $K{=}5$};
      \node[priv, text width=1.7cm] (ztrain) at (7.25,-4.0) {training:\\ one-hot $\mathbf{e}_{y_t}$};
      \node[sensed, text width=2.0cm] (zdeploy) at (7.25,-5.25)
      {deployment: hard argmax / soft softmax};
      \node[sensed, text width=1.3cm] (zt) at (9.5,-5.25) {Routing $\mathbf{z}_t$};
      \node[sensed, text width=1.9cm] (obs) at (11.5,-4.0) {Proprioception $\mathbf{o}_t$};
      \node[learned, text width=2.0cm] (experts) at (11.5,-5.25)
      {Experts $\pi_{\theta_1},\dots,\pi_{\theta_5}$};

      \node[sensed, text width=1.9cm] (rasters) at (3.0,-7.5)
      {Rasters $Z,E,Q$};
      \node[sensed, text width=1.5cm] (comstate) at (1.0,-8.35) {CoM state, duration $T$};
      \node[sensed, text width=1.5cm] (dcm) at (3.0,-8.35) {DCM $\boldsymbol{\xi}_T$};
      \node[sensed, text width=1.5cm] (velcmd) at (1.0,-9.25) {Velocity command};
      \node[sensed, text width=1.5cm] (pnom) at (3.0,-9.25) {Stride $\mathbf{p}_{\mathrm{nom}}$};
      \node[learned, text width=1.9cm] (cost) at (5.25,-8.35)
      {Cost field $\mathcal{J}(\mathbf{x})$ + DCM + E + Q\\
        + step-height M + climb bonus + window pen.};
      \node[learned, text width=1.6cm] (gradsteps) at (7.35,-8.35) {$N{=}6$ closed-form\\ gradient-descent steps};
      \node[outnode, text width=1.5cm] (qf) at (9.25,-8.5) {Refined foothold $\mathbf{q}_f$};
      \node[sensed, text width=1.5cm] (p0) at (9.25,-7.5) {Lift-off $\mathbf{p}_0$};
      \node[outnode, text width=2.0cm] (bezier) at (11.5,-8.03)
      {Adaptive quadratic B\'ezier swing $\mathbf{p}_s(t)$};

      \node[outnode, text width=3.0cm] (actor) at (15.25,-4.0)
      {Actor\\ $\mathbf{u}_t=\sum_k z_{t,k}\pi_{\theta_k}(\mathbf{o}_t)$};
      \node[sensed, inner sep=2pt] (robot) at (15.25,-7.25) {Deployed robot\\
        \includegraphics[width=4.5cm]{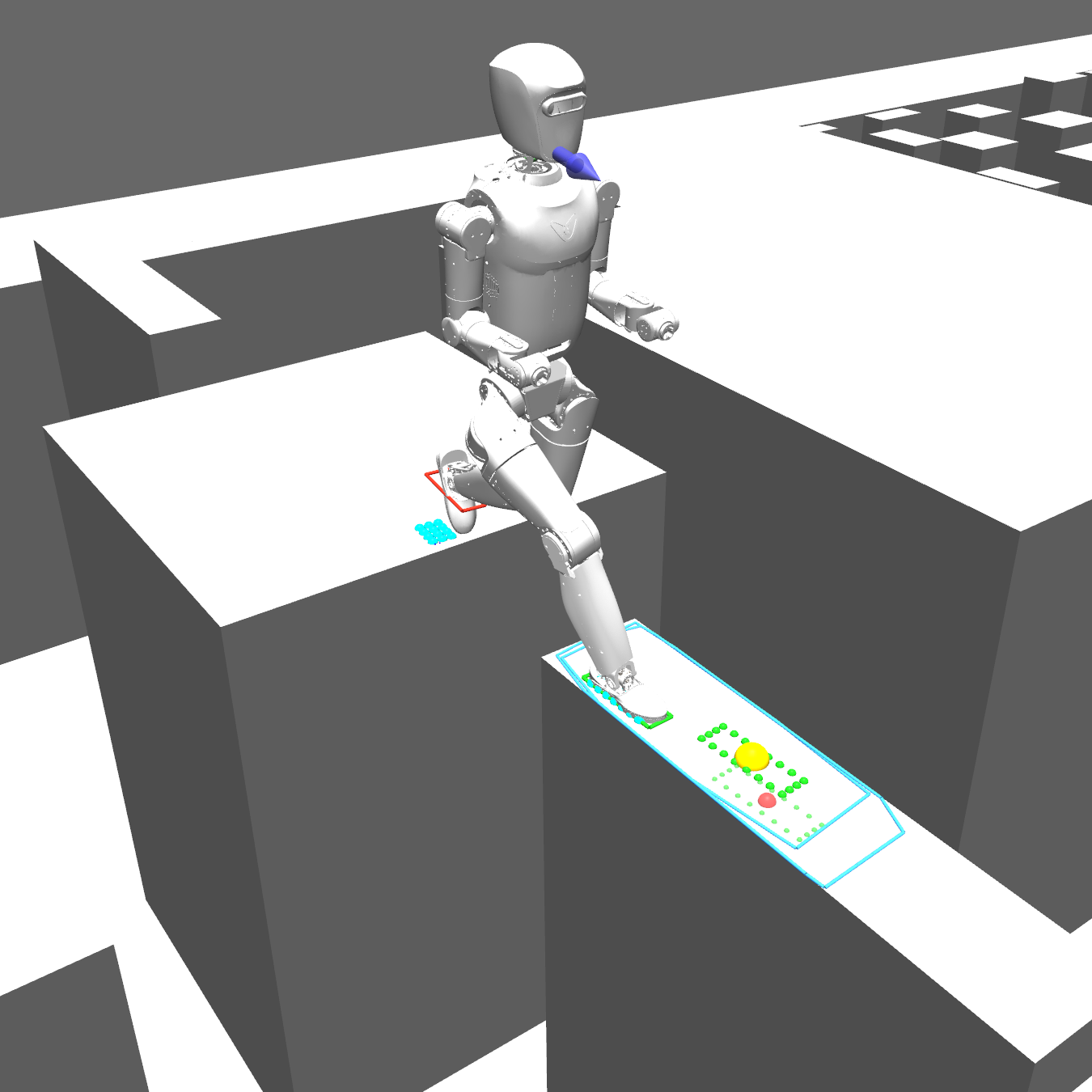}};

      \draw[dashed, line width=1pt, gray!60!black] (0,-3.15) -- (17.5,-3.15);

      \draw[arr] (depth.east) -- (enc.west);
      \draw[arr] (depth.south east) to[bend right=10] (ylabel.west);
      \draw[arr] (enc.east) -- (at.west);
      \draw[oarr] (at.east) -- node[elab, pos=0.55] {Config A} (critsig.west);
      \draw[oarr] (ylabel.east) -- node[elab, pos=0.5] {Config B} (critsig.west |- ylabel.east);
      \draw[arr] (critsig.east) to[bend right=35] (critic.west);
      \draw[oarr] (critpriv.south) -- (critic.north);
      \draw[arr] (ylabel.south west) to[bend right=25] node[elab, pos=0.7] {supervises\\(cross-entropy)} (gate.north);

      \draw[arr] (elev.east) -- (gate.west);
      \draw[arr] (gate.east) -- (logits.west);
      \draw[arr] (logits.north) to[bend left=25] (ztrain.west);
      \draw[arr] (logits.east) -- (zdeploy.west);
      \draw[oarr] (ztrain.east) to[bend left=35] node[elab, pos=0.65] {oracle route} (zt.north);
      \draw[arr] (zdeploy.east) -- (zt.west);
      \draw[arr] (zt.east) -- (experts.west);
      \draw[arr] (obs.south) -- (experts.north);
      \draw[arr] (elev.south) to[bend right=35] node[elab, pos=0.7] {shared} (rasters.west);

      \draw[arr] (comstate.east) -- (dcm.west);
      \draw[arr] (velcmd.east) -- (pnom.west);
      \draw[arr] (rasters.east) -- (cost.west |- rasters.east);
      \draw[arr] (dcm.east) -- (cost.west);
      \draw[arr] (pnom.east) -- (cost.west |- pnom.east);
      \draw[arr] (cost.east) -- (gradsteps.west);
      \draw[arr] (gradsteps.east) -- (qf.west);
      \draw[arr] (qf.east) -- (bezier.west |- qf.east);
      \draw[arr] (p0.east) -- (bezier.west |- p0.east);

      \draw[arr] (experts.east) -- (actor.west);
      \draw[arr] (bezier.north east) to[bend left=25] (actor.south west);
      \draw[arr] (actor.south) -- (robot.north);
    \end{tikzpicture}%
  }
  \caption{\textbf{The deployment boundary.} Solid blue/green/teal runs unchanged in both
    simulators and on a robot; dashed orange is privileged and simulation-only; dashed grey is
    frozen offline. The privileged label $y_t$ supervises the gate by cross-entropy and, in
    Configuration~B, conditions the critic; $a_t$ is the self-supervised alternative conditioner
    compared against it in Configuration~A. The deployed actor consumes neither $a_t$ nor $y_t$;
    the Bézier swing output reaches the actor directly, not through proprioception. Double-anonymous:
    no lab, codename, or affiliation identifiers appear on the figure.}%
  \label{fig:arch}
  \vspace{-3mm}
\end{figure*}

Contemporary state-of-the-art perceptive humanoid locomotion increasingly depends on curated,
terrain-conditioned motion-prior data, which is obtained from retargeted motion capture feeding a generative
motion-prior model, then distilled into a deployable
policy as done in ~\cite{zhang2026learningwholebodyhumanoidlocomotion,guo2026tgmpterrainconditionedgenerativemotion}.
We show that pipeline is not necessary. A Mixture-of-Experts policy gated by a terrain
classifier stands in for the motion-prior encoder, and a model-based foothold planner for
motion-capture-derived contact behaviour. This has proven to complete a
five-segment waypoint course end-to-end in $18$ of $20$ attempts and every
individual segment at $100\%$ on four of five, evaluated through the full
deployment stack in a second, deployment-fidelity simulator.
\Cref{fig:arch} shows how the parts connect, and which
exist only at training time.

The contributions of this system paper are:
\begin{enumerate}[leftmargin=1.25em,itemsep=1pt,topsep=1pt]
  \item \textbf{A locomotion system with no motion-prior dependency, and its transfer result.} The
        only privileged signal in the system is a scalar terrain class, distilled into a
        gate. Through the full deployment stack in a second simulator it crosses a challenging
        terrain course, $90\%$ of the time end-to-end.
  \item \textbf{Continuous foothold refinement in place of a discrete grid search, with the
          foothold target coming from a model-based approach.}
        Closed-form-gradient steps on a bilinearly interpolated terrain-cost field, batched
        across parallel environments every control step, lower landing error on three of four terrain
        classes at $6.7\times$ the matched discrete planner's compute.
        Against a matched discrete-grid control, this produces $72.4\%$ sub-grid correction.
  \item \textbf{A self-supervised terrain representation and a component study of what does
          and does not account for the transfer.}
        A temporal-contrastive embedding $a_t\in\R^{64}$, trained on the robot's own depth
        stream with no terrain label, reaches $59.9\%$ linear-probe accuracy for terrain
        identity, above the alternatives evaluated under the same protocol; it conditions
        the critic, not the actor's gate, and does not drive Item 1's result. Under a
        re-scoring protocol with a {random-hard} control (\cref{sec:protocol}), forcing the
        gate's own top-1 expert is free ($+1.8\%$) while forcing a random expert every step
        costs $61.1\%$ of return, a within-run contrast where the checkpoint is its own
        control.
\end{enumerate}

\section{Related Work}\label{sec:related}

\noindent\textbf{Terrain-conditioned mixtures of experts.}
Mode-adaptive gating over expert networks predates the recent terrain-conditioned systems by
nearly a decade~\cite{zhang2018modeadaptive}, blending expert weights from limb-velocity state so
that gait phase need never be labelled, and with no terrain label of any kind.
CMoE~\cite{ma2026cmoecontrastivemixture} clusters terrain with a SwAV-style contrastive objective
behind a soft gate, and reports the same failure mode this paper measures: a vanilla gate
collapsing to near-uniform expert activation. TAGA~\cite{li2026tagaterrainawareactivegaze} reuses
that router behind an emergent, RL-only active-gaze crop; CoRe-MoE~\cite{huang2026coremoe} is a
third soft, label-free system in the family; and CTS-MoE~\cite{affonso2026ctsmoe} is the nearest
neighbour to the architecture here, pairing a dense expert mixture under perception-based gating
with per-task critic heads, again without a terrain label. CMoE and TAGA are hardware-validated at
$80$--$120$\,cm gap crossing, and none of the four uses a privileged label. Our label-free router
is architecturally the vanilla gate their papers report collapsing, using cosine-similarity
logits, a temperature anneal, and a Switch-Transformer balance loss~\cite{fedus2022switch}, and
not their remedy for it. We also reimplemented CMoE's own SwAV/Sinkhorn--Knopp router and found it
does not rescue specialization on our task, a matched-task test and not a refutation, since
curriculum, policy strength and training details all differ. Turrisi et al.~\cite{turrisi2026moerl} show externally conditioned
expert selection beating monolithic policies, without a learned-router baseline. General practice
routes to a learned expert subset with a soft, top-$k$ combination and a load-balancing
term~\cite{shazeer2017moe,fedus2022switch}, with discrete gates relaxed via a straight-through
estimator~\cite{bengio2013estimatingpropagatinggradientsstochastic}.

\noindent\textbf{Self-supervised state and terrain representation.}
Contrastive predictive coding~\cite{oord2018cpc} and time-contrastive
networks~\cite{sermanet2018tcn} learn from temporal structure alone, and ATC~\cite{stooke2021atc}
fits such a representation offline for frozen downstream control. Our encoder applies that
construction to a legged robot's depth stream, using the parallel environment as the discriminated
instance, as CMoE does over gate logits. Depth-distillation pipelines decouple perception and
control for zero-shot
transfer~\cite{agarwal2022leggedlocomotionchallengingterrains,miki2022wild}, and follow-on work
reconstructs egocentric height maps~\cite{song2025gaitadaptive}, trains transformer policies on
outdoor terrain~\cite{radosavovic2024learninghumanoidlocomotionchallenging}, climbs stairs from
LiDAR elevation~\cite{long2024learninghumanoidlocomotionperceptive}, and extends to voxel and
multi-view representations~\cite{ben2025gallant,sun2026nowyousee}. A separate line replaces expert
structure with generative motion priors trained on retargeted motion
capture~\cite{zhang2026learningwholebodyhumanoidlocomotion,guo2026tgmpterrainconditionedgenerativemotion};
the representation and planner here consume neither, and the two routes are not compared under a
shared benchmark. Two recent depth-conditioned systems,
CReF~\cite{hao2026crefcrossmodalrecurrentfusion} and DPL~\cite{sun2025dpldepthonlyperceptive},
transfer to hardware, which we do not; neither reports the leakage-controlled representation
metric or the forced-routing protocol used here, so the two lines measure different things.

\noindent\textbf{Foothold planning.}
Model-based planners select landing targets under capturability constraints: MIQP over convex
regions~\cite{acosta2025perceptive}, joint step-timing and placement under DCM
bounds~\cite{xiang2026perceptivetiming}, RL-trained footstep policies over elevation
maps~\cite{kim2025visionfootstep}, and GPU-parallel discrete DCM search as an RL
reward~\cite{liu2026faststair}. ARTO-AL~\cite{wang2023artoal} re-plans location and timing jointly
online; our planner refines location only, at a fixed cadence, and is not compared against
timing-adaptive or mixed-integer planners empirically. Region-based mixed-integer methods already
optimize continuously inside a selected discrete
region~\cite{acosta2025perceptive,xiang2026perceptivetiming}, and
TAMOLS~\cite{jenelten2022tamols} optimizes base pose and footholds jointly against a raw heightmap
in under $10$\,ms by graduated optimization over progressively smoothed maps, the standard remedy
for the local minima an interpolated cost field also has. Gradient descent on an interpolated cost
field is CHOMP's construction~\cite{ratliff2009chomp}, reported before for legged foothold
selection over an interpolated heightmap costmap~\cite{yin2023footholds}.
BeamDojo~\cite{wang2025beamdojo} learns a sparse foothold reward directly, and
DTC~\cite{jenelten2024dtc} keeps an online model-based planner at deployment and replaces the
whole-body tracking controller with a learned policy, the converse of the split used here. Our
planner's scope is narrow: a batched local refinement inside a GPU-parallel rollout, against the
discrete-grid planner it replaces.

\section{System}\label{sec:method}

\subsection{Exteroceptive Sensing, and the Two Simulators}\label{sec:sensing}

Neither the gate (\cref{sec:gate}) nor the foothold optimizer (\cref{sec:foothold}) consumes
privileged simulation state. Both operate on a sensed local elevation grid
$\mathcal{H}_t\in\R^{H\times W}$ of world-frame ray hits registered into the robot's local
frame~\cite{miki2022wild,pmlr-v164-rudin22a}, produced by a simulated ray-caster at training time
and by a depth-to-elevation registration pipeline at deployment. The patch spans
$1.2\,\mathrm{m}\times1.0\,\mathrm{m}$ at $5\,\mathrm{cm}$ resolution, and control runs at
$50\,\mathrm{Hz}$. Training and deployment setups differences are noted in \cref{tab:stacks}.
The simulated ray-caster is noised (beam-angle noise, $\sigma=0.03^\circ$), which is not a
model of registration error, state-estimation drift, occlusion or latency.

\begin{table}[t]
  \centering
  \caption{Training setup against deployment setup.}
  \footnotesize
  \setlength{\tabcolsep}{3.5pt}
  \begin{tabular}{@{}lll@{}}
    \toprule
                     & Training                                               & Deployment                    \\
    \midrule
    Policy, rate     & \multicolumn{2}{c}{identical exported graph, $50$\,Hz}                                 \\
    Elevation source & simulated ray-caster                                   & depth-to-elevation pipeline   \\
    Camera           & $30$\,Hz ray-cast depth                                & $30$\,Hz, jitter logged       \\
    Dynamics         & \texttt{mjlab}/MuJoCo                                  & deployment-fidelity simulator \\
    Transport, log   & in-process tensors                                     & middleware stack, rosbag      \\
    \bottomrule
  \end{tabular}
  \label{tab:stacks}
  \vspace{-5mm}
\end{table}

\subsection{Expert Mixture and Gate}\label{sec:gate}

A small convolutional encoder $g_\phi$ maps $\mathcal{H}_t$ to logits over $K$ experts,
\begin{equation}
  \boldsymbol{\ell}_t = g_\phi(\mathcal{H}_t)\in\R^{K},\qquad
  q_{\phi,k} = \frac{\exp(\ell_{t,k})}{\sum_j\exp(\ell_{t,j})},
  \label{eq:logits}
\end{equation}
with $K{=}5$ fixed at construction from the terrain-class taxonomy of the training curriculum:
rough, stair-up, stair-down, gap, and hurdles. Expert $k$ is pre-assigned to class $k$ and is not
discovered. $K$ is tied to the taxonomy because the gate is trained by direct cross-entropy
against the label, unlike the label-free routers of \cref{sec:related} where $K$ is free. Let
$y_t$ denote the privileged, simulation-only terrain-class label, available from the robot's
position relative to the known terrain layout and unavailable on hardware. $g_\phi$ trains by
cross-entropy with its own optimizer, detached from the policy gradient:
\begin{equation}
  \mathcal{L}_{\mathrm{gate}}(\phi) = -\mathbb{E}_{(\mathcal{H},y)\sim\mathcal{D}}
  \big[\log q_{\phi,y}(\mathcal{H})\big].
  \label{eq:gateloss}
\end{equation}
Training routes on the ground-truth one-hot, and only deployment routes on the classifier's own
prediction:
\begin{equation}
  \mathbf{z}_t =
  \begin{cases}
    \mathbf{e}_{y_t},       & \text{policy training and oracle evaluation},       \\
    \widehat{\mathbf{z}}_t, & \text{deployment and predicted-routing evaluation},
  \end{cases}
  \label{eq:regime}
\end{equation}
with $\widehat{\mathbf{z}}^{\,\mathrm{hard}}_t=\mathbf{e}_{\argmax_k \ell_{t,k}}$,
$\widehat{z}^{\,\mathrm{soft}}_{t,k}=\mathrm{softmax}(\boldsymbol{\ell}_t/\tau)_k$, or an
adaptive interpolation between the two, mixing toward the soft vector in proportion to the
disagreement of $M{=}8$ Monte-Carlo-dropout passes. The experts therefore never experience gate
misclassification during training, and \cref{sec:protocol} measures that gap instead of absorbing
it. The trained gate reaches $97.1\%$ accuracy against the held-out label, and $97.2\%$ if
the $10\%$ least confident decisions are rejected. The action is the routed mixture,
\begin{equation}
  \mathbf{u}_t = \sum\nolimits_{k=1}^{K} z_{t,k}\,\pi_{\theta_k}(\mathbf{o}_t).
  \label{eq:mixture}
\end{equation}
No policy gradient reaches $g_\phi$: during training $\mathbf{z}_t=\mathbf{e}_{y_t}$ is
independent of $\phi$, and $\phi$'s optimizer is detached in any case.

For the label-free comparison we train an architecturally distinct router end-to-end with one PPO
objective plus routing auxiliaries: cosine-similarity logits, a temperature anneal
$\tau: 2.0\to0.7$, and the load-balancing loss of the Switch
Transformer~\cite{fedus2022switch}, which is often conflated with the
coefficient-of-variation importance loss of~\cite{shazeer2017moe},
\begin{equation}
  \mathcal{L}_{\mathrm{balance}} = K\sum\nolimits_{k=1}^{K}\sg(\bar f_k)\,\bar P_k,
  \label{eq:balance}
\end{equation}
where $\bar f_k$ is the batch fraction whose top-1 assignment is expert $k$, $\bar P_k$ the mean
soft mass on $k$, and $\sg$ stop-gradient.

\subsection{Continuous Local Foothold Refinement}\label{sec:foothold}

We use the DCM and capture-point framework~\cite{pratt2006capturepoint,englsberger2015dcm} with
$\omega_0=\sqrt{g/h_{\mathrm{CoM}}}$ and the linear-CoM-height reduction
of~\cite{liu2026faststair}. Write
$\boldsymbol{\xi}=\mathbf{c}_{\mathrm{CoM}}+\dot{\mathbf{c}}_{\mathrm{CoM}}/\omega_0$ for the
divergent component of motion, $T$ for the remaining step duration, $\boldsymbol{\xi}_T$ for its
predicted value at touchdown and $\mathbf{b}_{\mathrm{nom}}$ for the nominal DCM offset. The
nominal stride target is
$\mathbf{p}_{\mathrm{nom}}=(v_xT,\,v_yT+\mathrm{sign}(f)l_p)$ with $l_p$ the nominal lateral
stance width~\cite{khadiv2020steptiming,koolen2012capturability}. The cost channels below, the
swing arc, and the discrete planner of Configuration H are the multi-channel formulation
of~\cite{ly2026tactfulmultichannelterrainaffordance}; what is new here is the continuous
refinement that replaces its grid search.

Instead of scoring a discrete set of grid points, the terrain metrics a discrete planner already
computes, height $Z$, footprint-max-pooled Sobel gradient magnitude $E_{\mathrm{grid}}$ for
steepness, and footprint height range $Q_{\mathrm{grid}}$, become continuously queryable by
bilinear interpolation $\mathcal{I}[\cdot]$ at any $\mathbf{x}=(x,y)$. $E$ and $Q$ interpolate
precomputed rasters and are not analytic functions of $z(\mathbf{x})$; in particular
$E\neq\|\nabla z\|$, and the flatness channel penalizes an inclined plane even when perfectly
planar. Writing $[r]_+=\max(r,0)$ and $\delta z(\mathbf{x})=z(\mathbf{x})-z(\mathbf{p}_\ell)$
against the stance foot $\mathbf{p}_\ell$,
\begin{align}
   & z             = \mathcal{I}[Z](\mathbf{x}), \quad
  E     = \mathcal{I}[E_{\mathrm{grid}}](\mathbf{x}), \quad
  Q     = \mathcal{I}[Q_{\mathrm{grid}}](\mathbf{x}),                              \nonumber \\
   & M(\mathbf{x}) = \big[\,|\delta z(\mathbf{x})| - h_{\mathrm{eff}}\,\big]_+^2 ,
  \label{eq:M}
\end{align}
with $h_{\mathrm{eff}}$ the velocity-aware reachable step height. The full cost adds the
capture-point term, an anisotropic position term matching the discrete planner it replaces
($\lambda_y>1$ penalizes lateral deviation more strongly), a climb bonus
$b(\mathbf{x})=\mathbf{1}[v_x>v_{\mathrm{th}}]\min([\delta z]_+,h_{\mathrm{eff}})$, and a
squared-hinge penalty on leaving a $0.6\,\mathrm{m}$ square window centred at $\mathbf{x}_c$,
$W(\mathbf{x})=\sum_{d}[\,|x_d-x_{c,d}|-r_d\,]_+^2$:
\begin{align}
  \mathcal{J}(\mathbf{x}) & =
  \alpha_{\mathrm{pos}}\big[(x{-}p_{\mathrm{nom},x})^2 + \lambda_y (y{-}p_{\mathrm{nom},y})^2\big]
  \nonumber                                                                                                              \\
                          & \quad + \alpha_{\mathrm{dcm}}\|\mathbf{x}-(\boldsymbol{\xi}_T-\mathbf{b}_{\mathrm{nom}})\|^2
  + \alpha_E E + \alpha_Q Q \nonumber                                                                                    \\
                          & \quad + \alpha_M M - \alpha_{\mathrm{climb}} b + \alpha_{\mathrm{window}} W .
  \label{eq:cost}
\end{align}
$\mathcal{J}$ is continuous and piecewise differentiable in $\mathbf{x}$: the squared hinges
$M$ and $W$ are $C^1$ at their own thresholds, so the only first-derivative jumps come from the
grid-square boundaries of the bilinear interpolant and from the climb bonus, which is clipped
linear in $\delta z$. $\mathcal{J}$ is not convex, since two flat treads either side of a gap
produce two basins, so \cref{eq:refine} carries no global-optimality guarantee.

The target follows from $N$ explicit first-order steps out of the nominal stride target,
\begin{equation}
  \mathbf{x}^{(0)}=\mathbf{p}_{\mathrm{nom}},\quad
  \mathbf{x}^{(n+1)}=\mathbf{x}^{(n)}-\gamma\,\nabla_{\mathbf{x}}\mathcal{J}(\mathbf{x}^{(n)}),\;\;
  \mathbf{q}_f=\mathbf{x}^{(N)},
  \label{eq:refine}
\end{equation}
with $N{=}6$, fixed step size $\gamma$, and $\nabla_{\mathbf{x}}\mathcal{J}$ in closed form inside
a grid square. Retaining only the current iterate, memory is constant in $N$ and runtime linear.
$\boldsymbol{\alpha}$ is fixed and no gradient crosses \cref{eq:refine} in either direction. The
returned point is an approximate local refinement of an already-reasonable candidate, obtained
cheaply inside a GPU-batched rollout, a regime mixed-integer branch-and-bound does not target at
this scale.

The swing foot tracks a quadratic B\'ezier arc from lift-off $\mathbf{p}_0$ through control point
$\mathbf{c}$ to $\mathbf{p}_f=[\mathbf{q}_f;\,z(\mathbf{q}_f)]$, with a tangent-derived
orientation reference, evaluated on the continuous height field instead of at a grid point:
\begin{equation}
  \mathbf{p}(s)=(1{-}s)^2\mathbf{p}_0+2(1{-}s)s\,\mathbf{c}+s^2\mathbf{p}_f,\quad s\in[0,1].
\end{equation}
$\mathbf{c}$ lies off the curve and is not the apex: for a target apex $z_A>\max(z_0,z_f)$ the
required control height is $c_z = z_A+\sqrt{(z_A-z_0)(z_A-z_f)}$. The horizontal control point is
biased toward the higher endpoint by
$\rho=\clip(\tfrac12+\kappa(z_f-z_0)/h_{\max},\rho_{\min},\rho_{\max})$.

\subsection{Terrain Representation, Critic, and Training}\label{sec:repr}

A separate encoder $f_\psi$ maps the actor's onboard depth stream to $a_t\in\R^{64}$ with no
terrain label in its loss. Let $v_i = f_\psi(o_i)$ for anchor frame $i$ from parallel environment
$e(i)$, let $p(i)$ be a frame from the same environment a few control steps later, and let
$\mathcal{C}(i)$ index the in-batch candidates, which contain the positive and exclude the anchor
itself. The objective is InfoNCE over that pairing:
\begin{equation}
  \mathcal{L}_{\mathrm{TC}} = -\frac{1}{|\mathcal{A}|}\sum_{i\in\mathcal{A}}
  \log\frac{\exp\!\big(\mathrm{sim}(v_i,v_{p(i)})/\tau_c\big)}
  {\sum_{j\in\mathcal{C}(i)}\exp\!\big(\mathrm{sim}(v_i,v_j)/\tau_c\big)},
  \label{eq:tc}
\end{equation}
with $\mathrm{sim}$ the cosine similarity and $\tau_c$ a fixed temperature. Pairs never cross an
episode reset, and negatives are drawn from other environments in the batch. The anchor and
its positive sit a small fraction of a gait cycle apart, so phase is largely conserved across a
positive pair and the construction does not exclude a phase shortcut a priori. The encoder is fit once, offline, on rollouts from a
converged, label-supervised policy: its loss sees no label, though its data distribution does.
Each environment holds one terrain class for the whole episode, so a pooled nearest-neighbour
search would just retrieve the anchor's own episode; \cref{sec:parts} excludes the query's own
episode and environment from every metric, with probe splits at episode granularity.

The critic is a single value head whose backbone features are FiLM-modulated by $a_t$, replacing
the $K$ terrain-indexed value heads adapted from~\cite{sun2026nowyousee}, or, in the configuration
evaluated in \cref{sec:transfer}, the one-hot terrain label itself. Every critic variant also
receives the usual privileged training-only signals: uncorrupted elevation, ground-truth contact
forces, and true base velocity. The deployed actor receives none of them. Training minimizes
PPO~\cite{schulman2017ppo} plus $\lambda_{\mathrm{gate}}\mathcal{L}_{\mathrm{gate}}$ with adaptive
KL scheduling, GAE, and the standard terrain and velocity curricula~\cite{pmlr-v164-rudin22a}. The
gait-adaptive frequency head is adopted from~\cite{song2025gaitadaptive}. Simulation runs in
\texttt{mjlab} over MuJoCo~\cite{todorov2012mujoco}.

\section{Evaluation Design and Protocol}\label{sec:protocol}

\begin{table}[t]
  \centering
  \caption{The seven trained configurations, at $20{,}000$ iterations and $1024$ environments.
    B isolates the critic's conditioning signal from A, architecture otherwise identical; its
    \cref{sec:transfer} checkpoint is trained further, so the transfer result and the $20$k
    comparisons are not the same checkpoint. All run the continuous planner except H.}
  \footnotesize
  \begin{tabular}{@{}llccc@{}}
    \toprule
    Config. & Gate                   & $K$ & Critic        & Foothold   \\
    \midrule
    A       & priv.\ label, adaptive & 5   & $a_t$-FiLM    & continuous \\
    B       & priv.\ label, adaptive & 5   & one-hot label & continuous \\
    D       & priv.\ label, entropy  & 5   & $a_t$-FiLM    & continuous \\
    E       & label-free, balanced   & 5   & --            & continuous \\
    F       & none ($K{=}1$)         & 1   & --            & continuous \\
    G       & priv.\ label, adaptive & 5   & indexed       & continuous \\
    H       & priv.\ label, adaptive & 5   & indexed       & discrete   \\
    \bottomrule
  \end{tabular}
  \label{tab:configs}
  \vspace{-2mm}
\end{table}

\textbf{Sim-to-sim course.} Configuration B is evaluated on a five-segment chained waypoint course
($20$ attempts) deployed through the same perception-to-actuation software a physical robot would
run, inside a second simulator built for deployment fidelity and not for training throughput, with
every channel rosbag-logged as it would be on hardware. Each segment is scored twice, alone and
spawned fresh at its start, and as a leg of the chained course, so a within-segment failure can be
told from one specific to arriving already in motion.

\textbf{Training-simulator course.} Configurations are also scored inside the training simulator
on a nine-segment straight-line course built from the curriculum's own terrain generators,
reported as normalized distance travelled before failure at three difficulty levels and under both
routing modes. That normalized distance and the sim-to-sim course's are different quantities and
are not compared here.

\textbf{Forced-routing protocol.} Finished checkpoints are re-scored over $256$ environments
$\times$ $3000$ steps in five regimes: \emph{trained} (the deployed predicted gate), \emph{top-1}
(forced hard argmax), \emph{random-hard} (a uniformly sampled expert redrawn every step),
\emph{uniform} (routing discarded), and \emph{oracle} (the ground-truth label, as in training).
Random-hard is the control for top-1: it preserves the one-hot form the experts trained under
while destroying the terrain correlation. It also destroys temporal persistence, switching
expert about $40$ times a second, which the experts never experienced, so it bounds the value of
the gate's choice without isolating it; a constant-per-episode random expert would separate the
two and is not run. Every label-supervised checkpoint is scored at both predicted and oracle
routing, so the train-to-deploy routing gap is measured.

\textbf{Metrics and their chance levels.} The natural defaults are wrong for this data. The
gate-concentration score of a row-normalized gate-mass matrix $G$ is
$S=\frac1C\sum_c\max_k G_{ck}$; $1/K=0.20$ is the smallest value it can take, not a chance level,
and $S{=}1$ is attainable when every terrain routes to one expert, so $S$ measures concentration
and not bijectivity. The $k$-NN label consistency is neighbourhood purity
$C_k=\frac{1}{Nk}\sum_i\sum_{j\in\mathcal{N}_k(i)}\mathbf{1}[y_i=y_j]$, whose chance level under
the measured class priors is $\sum_c p_c^2$, and probe chance is $\max_c p_c$; both exceed $1/K$
under the natural distribution, which is skewed ($\approx76\%$ rough). Every $k$-NN and probe
number here is therefore computed on a class-balanced, stratified subsample, putting chance at
$0.20$ up to the residual effect of grouping. A label-permutation null for the silhouette score
($1000$ shuffles per dataset, three seeds, geometry and class sizes fixed, one-sided
$p=(b{+}1)/1001$ with $b{=}0$ exceedances) puts the observed value $4$--$5$ standard deviations
below the null mean in every seed.

\textbf{Scope of the two evaluations.} \Cref{sec:transfer} scores the deployed system at
deployment fidelity and answers whether the system works; \cref{sec:parts} scores the training
distribution at a fixed $20$k budget and answers why it is built this way. Training places
each environment inside a single terrain type for a whole episode, so nothing there asks a policy
to switch behaviour mid-episode, and the chained course is the first setting that does.

\section{Sim-to-Sim Course Evaluation}\label{sec:transfer}

\subsection{Course Result}\label{sec:course}

This section reports Configuration B, with a privileged-label gate, a critic conditioned on the
one-hot label, and the continuous foothold planner (\cref{tab:configs}), deployed through the full
perception-to-actuation stack inside the second simulator with every channel rosbag-logged. The
course chains five segments: slope, height-field, randomly-spread obstacles, up/down stairs,
hurdles, in that order, each $\approx16$--$20$\,m, $84.6$\,m in total.
Segment names are terrain-generator names, not gate classes: slope and height-field fall under the
gate's \emph{rough} class, randomly-spread obstacles under \emph{gap}, and up/down stairs spans
\emph{stair-up} and \emph{stair-down}. No isolated gap segment appears, and the gate's
predicted-class distribution over the course was not logged.

\begin{figure}[htb]
  \centering
  \includegraphics[width=0.85\linewidth]{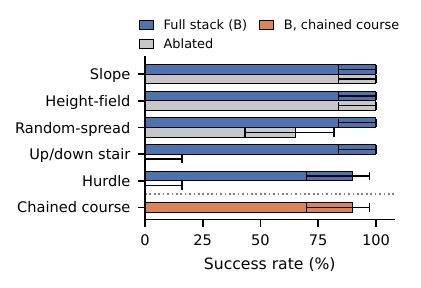}
  \caption{Sim-to-sim success, $n{=}20$ per bar, Wilson $95\%$ error bars. Each segment is
    scored alone; the chained course runs all five in order. The ablated configuration's own
    chained course is not shown, because it fails at hurdles and stairs terrain.}
  \label{fig:sim2sim}
  \vspace{-3mm}
\end{figure}

Scored alone, four segments succeed $20/20$ and the hurdle segment $18/20$
(\cref{fig:sim2sim,tab:nomoe}); scored as the chained course, end-to-end success is also $18/20$,
mean normalized distance $0.963\pm0.120$. At $n{=}20$ these carry Wilson $95\%$ intervals of
$[83.9,100]$ and $[69.9,97.2]$. They quantify evaluation uncertainty for one checkpoint.
Regarding the two chained failures, both are falls, not timeouts. One is at
$77.8$\,m, $10.3$\,m into the hurdle leg, and the two isolated hurdle failures are at $8.6$ and
$11.1$\,m into that same segment: the same failure mode at the same obstacle band whether the
segment is entered fresh or after $67$\,m of prior terrain. The other is at $41.2$\,m, $9.5$\,m
into randomly-spread obstacles, a segment that never fails in isolation. Chained success
therefore equals the weakest segment's isolated rate and does not compound across segments, and
the one failure per-segment testing misses is a cross-segment effect instead of a harder obstacle.

\subsection{Deployment-Stack Fidelity}\label{sec:fidelity}

Across the $20$ chained attempts
the perception pipeline holds $30$\,Hz to within $0.003$\,Hz; command-tracking RMSE is
$0.14$--$0.22$\,m/s by segment; $95$th-percentile peak ground reaction force is $2.78$--$3.16$
times half body weight; mean power runs $276$--$623$\,W. Per-foot six-axis force/torque, contact
and touchdown flags, commanded against actual body twist, and IMU roll/pitch are logged per
timestep at $20$\,Hz for the first three segments; median episode $95$th-percentile attitude
excursion is $6.5^\circ/4.0^\circ$ (roll/pitch) on slope, $7.6^\circ/3.2^\circ$ on height-field,
$7.0^\circ/3.3^\circ$ on randomly-spread obstacles.

\subsection{Comparison With an Ablated Configuration}\label{sec:ablated}

\begin{table}[t]
  \centering
  \caption{Sim-to-sim per-segment success with Wilson $95\%$ intervals, $n{=}20$ per
    segment; success in \%, Wilson interval subscripted, normalized distance in parentheses. The
    ablated configuration removes \emph{both} the mixture and the refinement and traverses
    $26$--$30\%$ faster. $p$: Fisher exact.}
  \footnotesize
  \setlength{\tabcolsep}{2.5pt}
  \begin{tabular}{@{}lccc@{}}
    \toprule
    Segment       & B                         & Ablated ($-$mix, $-$refine)   & $p$                \\
    \midrule
    Slope         & $100_{[84,100]}\,(0.977)$ & $100_{[84,100]}\,(0.977)$     & $1$                \\
    Height-field  & $100_{[84,100]}\,(0.977)$ & $100_{[84,100]}\,(0.977)$     & $1$                \\
    Random-spread & $100_{[84,100]}\,(0.977)$ & $65_{[43,82]}\,(0.761)$       & $8{\cdot}10^{-3}$  \\
    Up-down-stair & $100_{[84,100]}\,(0.982)$ & $\mathbf{_{[0,16]}}\,(0.324)$ & $1{\cdot}10^{-11}$ \\
    Hurdle        & $90_{[70,97]}\,(0.936)$   & $\mathbf{_{[0,16]}}\,(0.156)$ & $3{\cdot}10^{-9}$  \\
    \bottomrule
  \end{tabular}
  \label{tab:nomoe}
  \vspace{-2mm}
\end{table}

A configuration that removes both the expert mixture (a single network) and the continuous
refinement (discrete planner) was scored on the same five segments. It equals B on slope and
height-field, drops to $65\%$ on randomly-spread obstacles, and reaches $0\%$ on both
up-down-stair and hurdle, where B reaches $100\%$ and $90\%$: the gap is concentrated exactly on
the two terrain-specific segments ($p<10^{-8}$, Fisher) and absent on the two continuous ones
($p=1$). The failure shapes differ as much as the rates. On hurdles all $20$ episodes fall
between $2.6$ and $5.0$\,m, a hard stop and not a borderline; on stairs failures split evenly
between falls and getting stuck, over $0.6$--$18.8$\,m. \Cref{sec:planner}'s finding that
continuous refinement's advantage is itself concentrated on stairs makes the planner a plausible
contributor alongside routing; this comparison does not separate them.

Notwithstanding, on the two segments it does cross it is
the smoother walker by a wide margin: stumble rate $1.07$ and $0.79$ per metre against B's
$3.22$ and $2.67$, traversal time $24.7$\,s against $32.3$ and $31.0$\,s, and energy per metre
$287$ and $304$\,J/m against B's $751$ and $540$\,J/m. Read together with the table, the
mixture-plus-refinement system buys traversal capability on terrain a single generalist cannot
cross at all, and pays for it with roughly three times the stumble rate and twice the energy per
metre on terrain both can cross.

\section{Component Analysis}\label{sec:parts}

Every number here is on the training distribution at the fixed $20$k budget of
\cref{tab:configs}, and answers why the system is built this way, not how well it works.

\subsection{Foothold Planner}\label{sec:planner}

\begin{figure}[t]
  \centering
  \begin{subfigure}[b]{0.49\linewidth}
    \begin{overpic}[width=\linewidth,keepaspectratio]{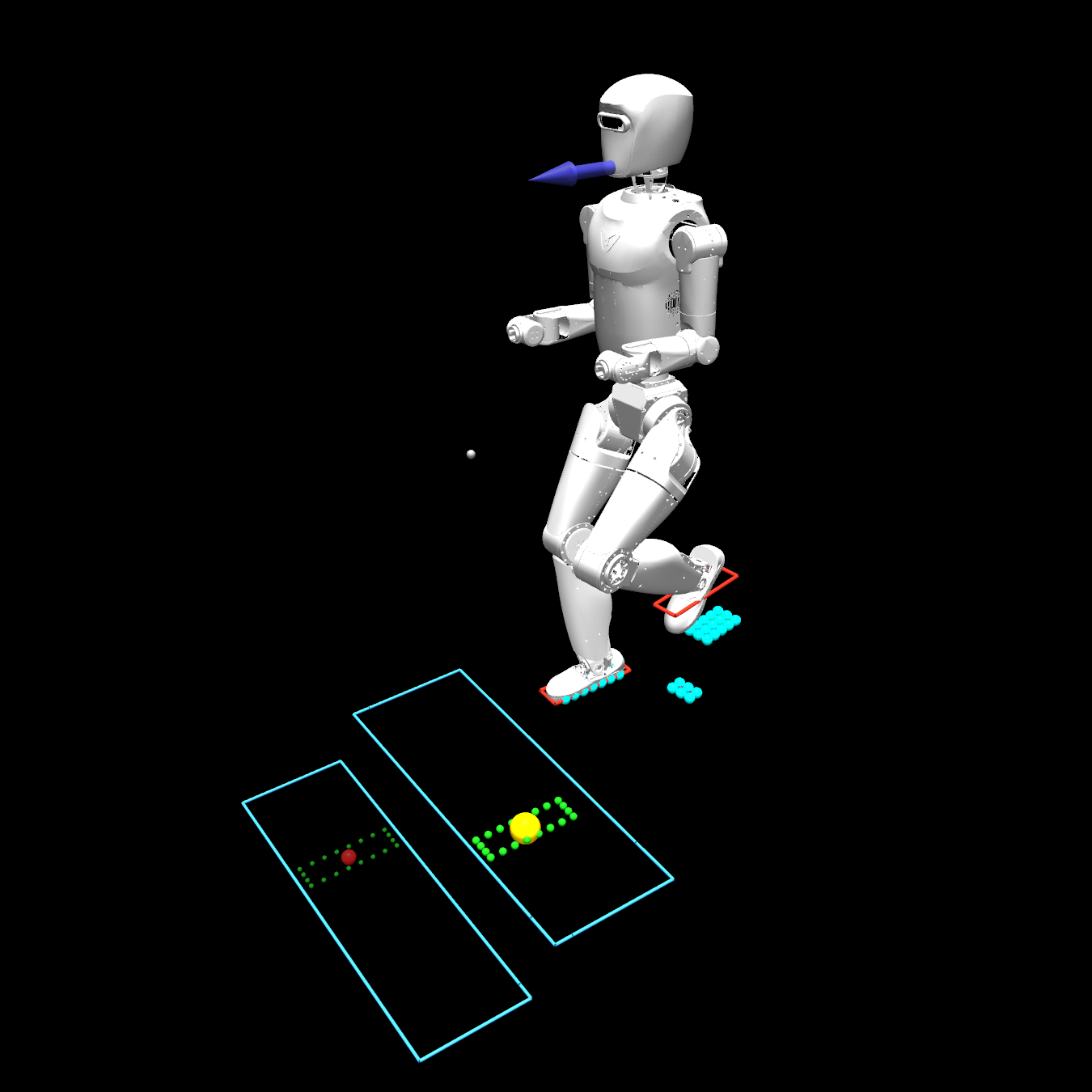}
      \put(3,3){\textcolor{white}{(a)}}
    \end{overpic}
  \end{subfigure}%
  \begin{subfigure}[b]{0.49\linewidth}
    \begin{overpic}[width=\linewidth,keepaspectratio]{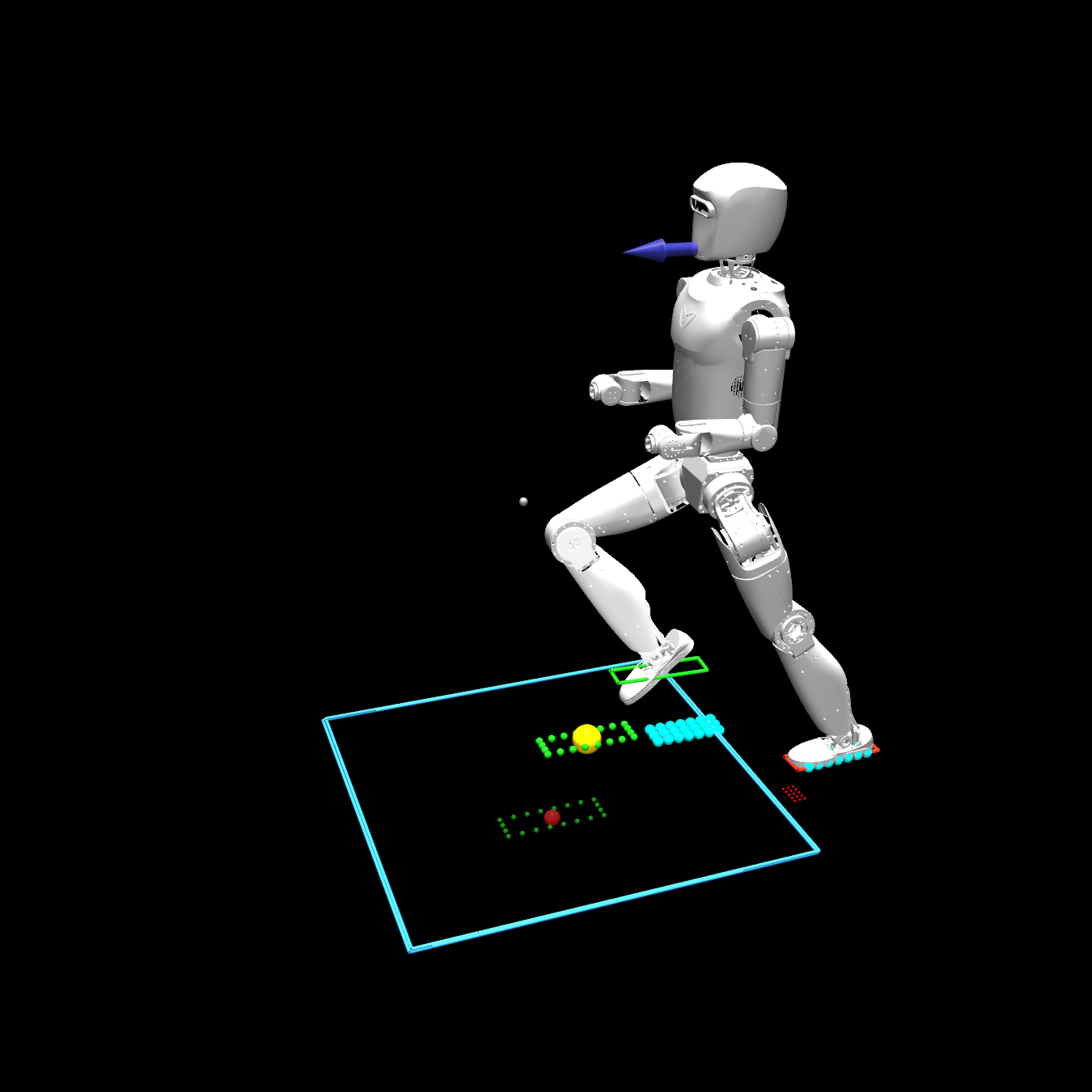}
      \put(3,3){\textcolor{white}{(b)}}
    \end{overpic}
  \end{subfigure}
  \begin{subfigure}[b]{0.49\linewidth}
    \begin{overpic}[width=\linewidth,keepaspectratio]{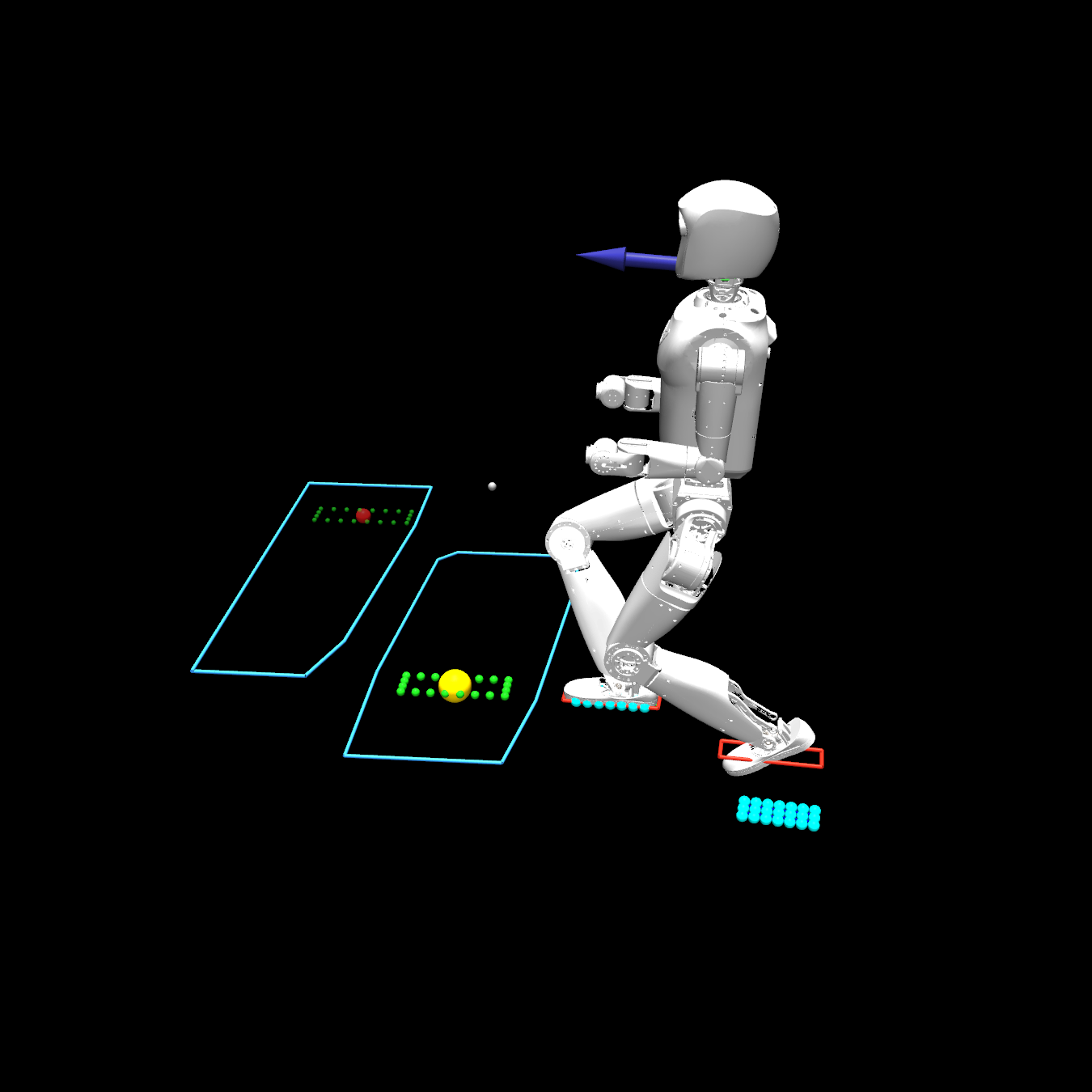}
      \put(3,3){\textcolor{white}{(c)}}
    \end{overpic}
  \end{subfigure}%
  \begin{subfigure}[b]{0.49\linewidth}
    \begin{overpic}[width=\linewidth,keepaspectratio]{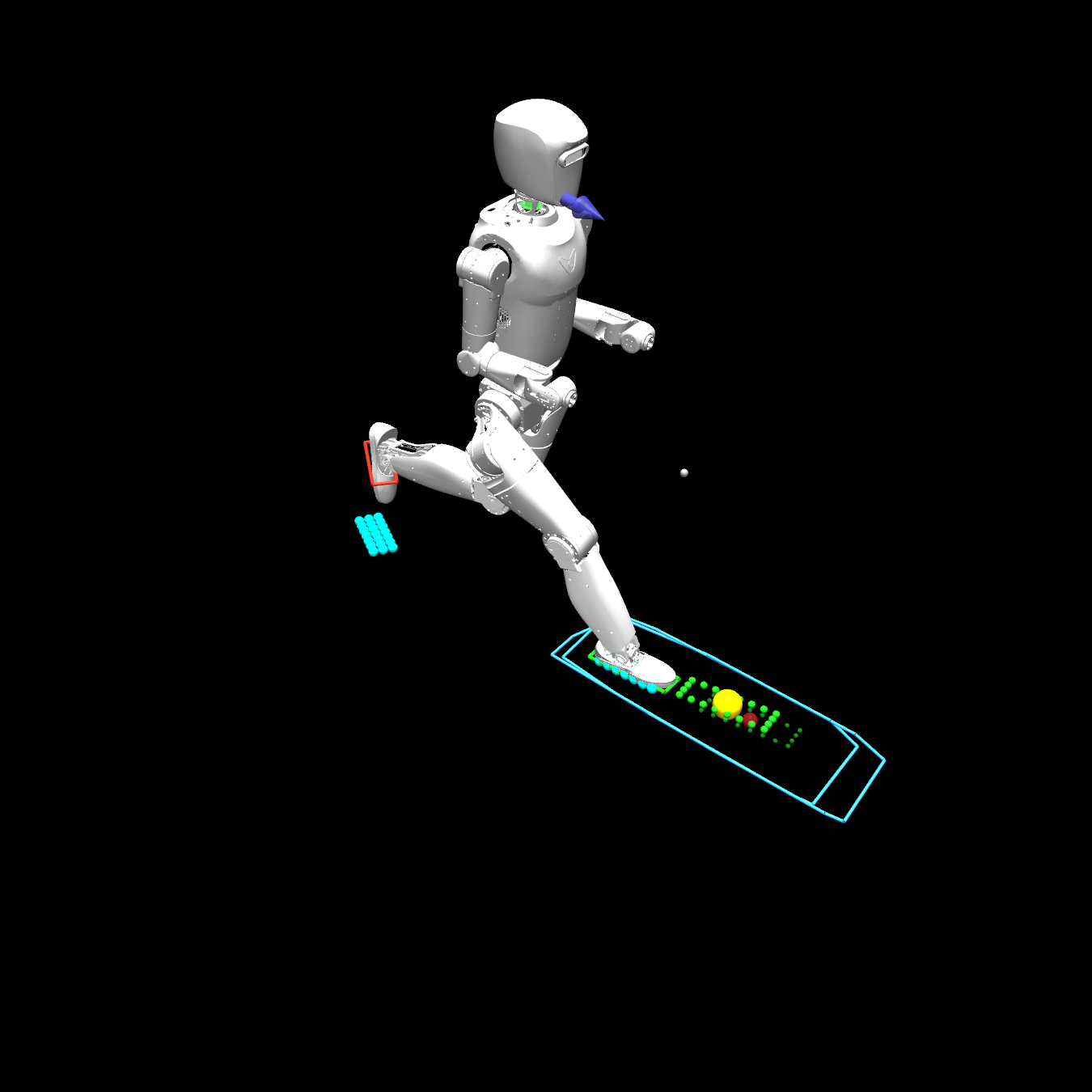}
      \put(3,3){\textcolor{white}{(d)}}
    \end{overpic}
  \end{subfigure}
  \caption{The continuous planner (G), terrain hidden so its debug markers are
    unoccluded. Bright marker: the swinging foot's refined target; dim marker: the stance foot's
    next planned landing spot; shaded rectangle: the region searched around the nominal stride
    target. (a)~stairs down, (b)~hurdles, (c)~stairs up, (d)~gap.}%
  \label{fig:footholdgallery}
  \vspace{-3mm}
\end{figure}

Configurations G and H share task, seed, and training budget, and differ in the foothold planner.
The continuous planner also replaces the discrete planner's hard search-window mask with the soft
penalty $W$ of \cref{eq:cost}, which the gradient needs to stay informative near the window edge.
Equalizing the two window treatments would isolate interpolation order, and is not run.

On the training distribution G is ahead of H on success at $5$\,m ($0.109\pm0.011$ against
$0.081\pm0.010$) and survival ($0.275$ against $0.265$), concentrated on stairs: per-terrain
success $0.15/0.29$ on stair-up/stair-down for G against $0.11/0.19$ for H. Landing error over
the training tail favours G on three of four classes, stair-up $0.107$ against $0.110$, stair-down
$0.023$ against $0.040$, hurdle $0.180$ against $0.246$, and favours H on gap ($0.104$ against
$0.117$).

\textbf{Compute cost.} The continuous planner costs $6.7\times$ the discrete planner's per-step
compute, $9.55$ against $1.42$\,ms per simulation step ($4.78$ against $0.71$\,ms per planner
call, two calls per step), over $500$ steps $\times$ $64$ environments.
It displaces $72.4\%$ of placements across at least one grid-square boundary, $74.8\%$ of
refinement trajectories visit more than one square, and mean seed-to-final displacement is
$1.33$ squares ($6.65$\,cm at the $5$\,cm grid), so the refinement is not a nudge inside one
square, and a hard per-square mask of the kind the discrete planner uses would clip roughly
three quarters of these moves. Crossing a boundary is a displacement statistic and not evidence
that the new placement is better; the landing error above is that evidence. These timings are
batched over $64$ environments while training runs $1024$: a throughput measurement, not
single-instance deployment latency, which is not measured here.

\textbf{Interpolation kernel.} A biquadratic ($3{\times}3$-tap) variant differing from G only in
interpolation order reaches a higher forced-routing return ($112.7$ against $104.5$ for G, both
from the separate scoring pass described in \cref{tab:routing}) and a larger top-1 margin
($+9.5\%$ against $+5.0\%$), but traverses less far on the training-simulator course at matched
budget ($0.086$ against $0.117$ normalized distance). Bilinear, used throughout, is not a
compromise choice.

\subsection{Routing}\label{sec:routing}

\begin{figure}[t]
  \centering
  \begin{subfigure}{.24\textwidth}
    \includegraphics[width=\linewidth]{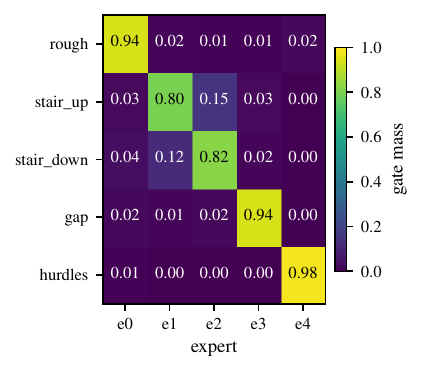}
    \caption{B (privileged label)}
  \end{subfigure}\hfill
  \begin{subfigure}{.24\textwidth}
    \includegraphics[width=\linewidth]{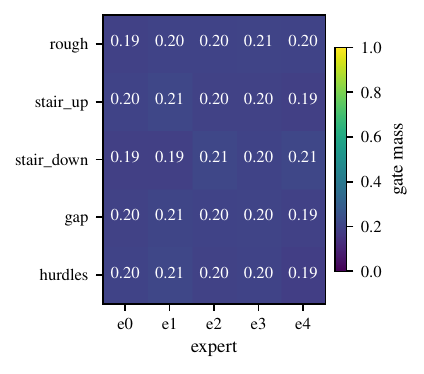}
    \caption{E (label-free)}
  \end{subfigure}
  \caption{Mean gate mass per terrain, row-normalized, on each final checkpoint. The
    privileged-label gate is near-permutation ($S{=}0.90$); the label-free router is
    indistinguishable from uniform ($S{=}0.21$, minimum $0.20$).}
  \label{fig:spec}
  \vspace{-5mm}
\end{figure}

\begin{table}[t]
  \centering
  \caption{Forced-routing re-scoring, $256$ environments $\times$ $3000$ steps. \emph{Trained} is
    the absolute mean return of the deployed gate; the other columns are percentage change against
    it, so a change past $-100$ takes the return below zero. \emph{Rand-hard} forces a uniformly
    sampled expert every step, the control for Top-1 (\cref{sec:protocol}). F ($K{=}1$) is
    omitted, since forced-hard and uniform routing are then no-ops. The Rand-hard column was
    collected in a separate scoring pass; rollouts are not bit-reproducible, so its returns
    differ by a few points from the other columns. Percentages use unrounded returns.}
  \footnotesize
  \setlength{\tabcolsep}{4pt}
  \begin{tabular}{@{}lccccc@{}}
    \toprule
    Config.   & Trained & Top-1           & Rand-hard        & Uniform          & Oracle \\
    \midrule
    A (priv.) & $97.7$  & $+3$            & $\mathbf{-55}$   & $\mathbf{-36}$   & $+9$   \\
    B (priv.) & $129.1$ & $+1.8$          & $\mathbf{-61.1}$ & $\mathbf{-34.0}$ & $+0.4$ \\
    D (priv.) & $96.2$  & $+7$            & $\mathbf{-57}$   & $\mathbf{-41}$   & $+13$  \\
    G (priv.) & $107.5$ & $+2$            & $\mathbf{-60}$   & $\mathbf{-39}$   & $+11$  \\
    E (free)  & $120.2$ & $\mathbf{-101}$ & $\mathbf{-121}$  & $-5$             & n/a    \\
    \bottomrule
  \end{tabular}
  \label{tab:routing}
  \vspace{-4mm}
\end{table}

\textbf{Interpreting the gate-concentration score.} A near-permutation gate-mass matrix follows
from construction as much as from learning: the gate is a classifier scored against the label it
trained on, experts are pre-assigned, and training routes on the ground-truth one-hot, so a
diagonal matrix is what an accurate classifier must produce. It does not imply that the $K$
experts became functionally distinct, since they could converge to one function and leave routing
inert, and discarding routing does not settle that either, because it conflates specialization
with brittleness to a gate distribution the experts never saw. Random-hard separates the two: if
the gate's top-1 choice is harmless while random-hard is catastrophic, the gate picks a
state-appropriate expert and not one member of an interchangeable set; if the two are
indistinguishable, that is an ensemble. \Cref{tab:routing} shows the first pattern for every
label-supervised configuration (top-1 $+1.8$ to $+7\%$, random-hard $-55$ to $-61\%$) and the
second for the label-free one.

\textbf{Four label-free routers behave alike.} The label-free router E sits at chance
concentration ($S{=}0.21$), and forcing its top-1 destroys the policy ($-101\%$). Three
further mechanisms behave the same way. Zeroing $\lambda_{\mathrm{balance}}$ raises
concentration to $0.335$ and makes uniform routing costlier ($-18\%$ against $-2\%$), yet leaves
top-1 and random-hard indistinguishable ($-108$ against $-123\%$). An actor-conditioned gate
routing directly on $a_t$ in place of any label reproduces E's signature exactly ($S{=}0.228$,
top-1 $-111\%$, random-hard $-128\%$, uniform $-6\%$). A reimplementation of CMoE's own
SwAV/Sinkhorn--Knopp router lands at $0.211$, matching the vanilla router to three decimals.
None of the four produces a router the policy uses as a router.
Cosine-similarity logits lie in $[-1,1]$, which caps the attainable maximum
soft mass at $r_{\max}\le[1+(K{-}1)e^{-2/\tau}]^{-1}\approx0.81$ for $K{=}5,\tau{=}0.7$.
That ceiling is far above the observed $0.21$ and does not by itself explain it; the
measured router logit standard deviation of $0.30$ does, capping the attainable maximum at
$0.31$ at the configured $\tau$ and $0.27$ at the $\tau{=}1.03$ the anneal actually reached.
The balance term contributes a second effect: with $\bar f$ detached,
$\partial\mathcal{L}_{\mathrm{balance}}/\partial\ell_{ik}\propto
  r_{ik}(\bar f_k-\sum_j \bar f_j r_{ij})$, which vanishes identically once dispatch is balanced,
whatever the per-sample sharpness, so the term applies no pressure toward sharp individual routing
once the aggregate is balanced. Balanced aggregate dispatch and uniform per-sample routing
are different conditions, and it is the second that holds here: logged per-sample gate entropy
settles at $1.58$ nats against the $K{=}5$ ceiling $\ln 5\approx1.61$, so the row-averaged
uniformity of \cref{fig:spec} is not an artifact of averaging over sharp, diverse per-sample
routes.

\subsection{Negative Results}\label{sec:negative}

\textbf{One expert equals five on the training distribution.} Configuration F is E's label-free
architecture collapsed to a single global expert ($K{=}1$). On the training
curriculum the six configurations at matched budget separate by about iteration $7000$ and finish
in the same $5$--$8$ band, F highest among them, while Configuration B, leaves the band and plateaus
near level $20$. F is also the \emph{smaller} model by $532$k
parameters ($1{,}307{,}994$ against $1{,}840{,}266$ for the $K{=}5$ label-free mixture; the
label-supervised gate adds a further $581$k, to $2{,}421{,}167$), so F equalling the mixtures
is not an artifact of a capacity handicap on F's side. A $K$ sweep is not meaningful on the
label-supervised family anyway, since $K$ there is the number of classes the gate classifies, not
a capacity knob.
This result and \cref{sec:transfer}'s sit in the two evaluations of \cref{sec:protocol}, and
the reading consistent with both is that routing is not what a single-terrain episode rewards:
training holds one terrain class per episode throughout, so a generalist never switches
behaviour mid-episode and a router has nothing to do, while on the chained course the
single-network configuration reaches $0\%$ on its two terrain-specific segments. This is the
reading the two results jointly support, not an established mechanism, since the ablated
configuration is not Configuration F and it changed the planner too.

\textbf{The uncertainty-adaptive gate hedge is a null result.} Configuration D (A with the
MC-dropout hedge disabled) tracks A within the spread of \cref{tab:routing}. The hedge's
calibration is a property of the gate's confidence against its own correctness, not of policy
rollouts: $10$-bin ECE $0.034$, AUROC $0.688$ over $n{=}384{,}000$ decisions, at $+3.5\%$ compute.
It flags incorrect routing better than chance and changes no policy outcome we measure.
\textbf{The self-supervised embedding loses to the label it replaces.} An architecture-matched
comparison put the label-conditioned critic ahead of the $a_t$-FiLM critic on every outcome:
success at $5$\,m $2.6\%\to15.3\%$, survival $21.7\%\to34.9\%$, fall rate $78.3\%\to65.1\%$,
stair-up $0\%\to18.8\%$, stair-down $2.7\%\to30.6\%$, gap $0\%\to15.4\%$ (hurdles $0\%$ for both).

\begin{table}[t]
  \centering
  \caption{Terrain-representation quality, cross-env: same-episode and same-environment
    neighbours excluded from $k$-NN, probe split grouped by environment. Mean $\pm$ spread over
    $n{=}3$ independent rollout datasets, on a class-balanced subsample.}
  \footnotesize
  \setlength{\tabcolsep}{3.5pt}
  \begin{tabular}{@{}lccc@{}}
    \toprule
    Objective                   & Probe                 & $10$-NN               & Silhouette     \\
    \midrule
    Temporal-contrastive (ours) & $\mathbf{59.9\pm2.0}$ & $\mathbf{59.5\pm1.8}$ & $-0.11\pm0.02$ \\
    Best of four alternatives   & $53.5\pm0.9$          & $47.1\pm2.2$          & --             \\
    Worst of four alternatives  & $35.8\pm3.1$          & $34.0\pm2.3$          & --             \\
    \midrule
    Chance                      & {$20.0$}              & {$20.0$}              & {perm.\ null}  \\
    \bottomrule
  \end{tabular}
  \label{tab:repr}
  \vspace{-2mm}
\end{table}

\textbf{The embedding does recover terrain structure without a label.} Four alternative
objective/target combinations, permuting $\{$reconstruction, contrastive$\}\times\{$contact,
swing-geometry$\}$, plateau at $35.8$--$53.5\%$ probe accuracy under the grouped protocol
(\cref{tab:repr}) for one diagnosable reason: a phase-prediction probe on the same frozen
embeddings (4-class gait phase from ground-truth contact, floor $48\%$) shows all four reach
$85$--$93\%$, a $+37$ to $+45$\,pp margin that dwarfs their $+16$ to $+34$\,pp terrain margin. The
temporal-contrastive objective reaches $59.9\%$, about $3\times$ chance, and the phase probe
inverts the ordering against it, $+40$\,pp of terrain margin against $+19$\,pp of phase margin,
the only one of the five where terrain beats phase relative to its own chance level. The
silhouette is negative because the classes are not compact convex clusters;
\cref{sec:protocol}'s permutation null rules out the reading that they are not separable at all,
though shuffling labels at point level does not exclude instance or temporal clustering as the
structure it detects. The terrain signal is in the robot's own exteroception, recoverable without a
label, before any router is asked to find it.

\section{Scope and Limitations}\label{sec:scope}

\textbf{No baseline from the competing line was run.} We compare against an ablated version of our
own system, not against a motion-prior system, and claim no superiority over that line of work
anywhere in this paper. The motion-prior systems we cite are closed-source; a reimplementation
from their published description did not reach reported performance, and we chose not to present
that as a strawman comparison.
\textbf{The transfer result is configuration-specific.} The gain is concentrated on the two
terrain-specific segments and is not uniform, and whether it comes from the mixture, the planner,
or both is not separated by the ablated configuration.
\textbf{Seeds.} Seed variance was large relative to most policy-outcome differences, and
\cref{sec:transfer} is one seed of Configuration B, so whether $90\%$ end-to-end is a stable
property of the configuration is not established here. Within-run comparisons, one checkpoint
re-scored under several routing regimes, remain their own control; between-run comparisons on the
training distribution do not.
\textbf{Training budget.} Every training-distribution and forced-routing number is scored at the
fixed $20{,}000$-iteration budget of \cref{tab:configs}, while \cref{sec:transfer}'s checkpoint is
trained further.
\textbf{Sim-to-sim is not hardware.} \Cref{sec:transfer} runs the full deployment software stack
against a second, deployment-fidelity simulator using rosbag, without any physical robot.
\textbf{Planner scope.} \Cref{eq:refine} enforces capturability as a soft cost, does not jointly
optimize step timing, has no global-optimality guarantee, and does not attempt disconnected-region
handling. Unlike the mixed-integer planners~\cite{acosta2025perceptive,xiang2026perceptivetiming}
it offers no certified safety guarantee. This is a scope choice with cheap local refinement inside a
GPU-batched rollout.

\section{Conclusion}\label{sec:conclusion}

A privileged-label Mixture-of-Experts policy with a continuously-refined model-based
foothold planner, carrying no motion-capture data and no learned motion prior, crosses a
five-segment waypoint course end-to-end $18$ of $20$ times under a sim-to-sim evaluation built
for deployment fidelity. The component study supports that the gate's choice of expert
contributes to this success ($-61\%$ of return when randomized, $+1.8\%$ when hardened), and the self-supervised
terrain embedding does not, measurably underperforming the privileged label wherever the
two have been substituted for each other on the critic. That
embedding remains a label-free result on its own terms, $59.9\%$ linear-probe terrain accuracy
against $35.8$--$53.5\%$ for four alternatives that collapse onto gait phase, so the terrain
signal is recoverable.

Generative-AI use. A large language model assisted with proofreading manuscript text,
cross-checking reported figures against logged experiment outputs, and producing
analysis and plotting scripts; all reported numbers were verified by the authors against the
source logs cited throughout this paper.

\bibliographystyle{IEEEtranBST/IEEEtran}
\bibliography{references}

\end{document}